\documentclass[conference]{IEEEtran}
\usepackage[utf8]{inputenc}
\usepackage{xcolor}
\usepackage{lipsum}  
\usepackage{booktabs}
\usepackage{hyperref} 
\usepackage{caption}
\usepackage{subcaption}
\usepackage{cite}
\usepackage{amsmath,amssymb,amsfonts}
\usepackage{algorithmic}
\usepackage{graphicx}
\usepackage{textcomp}
\usepackage{lipsum}  
\usepackage{tikz}

\usepackage{hyperref}
\usepackage{booktabs}
\usepackage{tabularx}
\usepackage{fancyhdr}

\newcommand{\dataset}{{\cal D}}

\newcommand{\realset}{{\mathbb{R}}}

\newcommand{\z}{\mathbf{z}}
\newcommand{\Z}{\mathbf{Z}}
\newcommand{\s}{\mathbf{s}}
\newcommand{\x}{\mathbf{x}}
\newcommand{\X}{\mathbf{X}}
\newcommand{\y}{\mathbf{y}}

\newcommand{\h}{\mathbf{h}}

\newcommand{\KL}{\text{KL}}

\newcommand{\phiX}{\varphi_{\omega}^{\text{\text{$\x$}}}}
\newcommand{\phiXs}{\varphi_{\omega}^{\text{\text{$\s$}}}}
\newcommand{\phiEnc}{\varphi_{\omega}^{\text{\text{enc}}}}
\newcommand{\phiDec}{\varphi_{\omega, d}^{\text{\text{dec}}}}
\newcommand{\phiPrior}{\varphi_{\omega}^{\text{\text{prior}}}}

\newcommand\copyrighttext{%
	\footnotesize \textcopyright 2021 IEEE. Personal use of this material is permitted.
	Permission from IEEE must be obtained for all other uses, in any current or future
	media, including reprinting/republishing this material for advertising or promotional
	purposes, creating new collective works, for resale or redistribution to servers or
	lists, or reuse of any copyrighted component of this work in other works.
	DOI: \href{https://doi.org/10.1109/JBHI.2021.3123839}{10.1109/JBHI.2021.3123839}}
\newcommand\copyrightnotice{%
	\begin{tikzpicture}[remember picture,overlay]
	\node[anchor=south,yshift=10pt] at (current page.south) {\fbox{\parbox{\dimexpr\textwidth-\fboxsep-\fboxrule\relax}{\copyrighttext}}};
	\end{tikzpicture}%
}
\begin{document}

%\title{Hiss-VAE: Heterogeneous Incomplete Sequential VAE}
\title{Medical data wrangling with sequential variational autoencoders}
\author{Daniel Barrej\'on, Pablo M. Olmos, Antonio Art\'es\text{-}Rodr\'iguez
\thanks{This work has been partly supported by Spanish government MCI under grants TEC2017-92552-
EXP and RTI2018-099655-B-100, by Comunidad de Madrid under grants IND2017/TIC-7618,
IND2018/TIC-9649, IND2020/TIC-17372 and Y2018/TCS-4705, by BBVA Foundation under the
Deep-DARWiN project, and by the European Union (FEDER) and the European Research Council
(ERC) through the European Union's Horizon 2020 research and innovation program under Grant
714161.}
\thanks{D. Barrej\'on, P.M.Olmos and A. Art\'es-Rodr\'iguez are with the Dept. of Signal Theory 
and Communications, UC3M, Legan\'es, Madrid, 28911. e-mails: (dbarrejon@tsc.uc3m.es, olmos@tsc.uc3m.es, antonio@tsc.uc3m.es). They are also with the Gregorio Mara\~n\'on Health Research Insitute (Spain).}
}

\maketitle
\copyrightnotice

\begin{abstract}

Medical data sets are usually corrupted by noise and missing data. These missing patterns are commonly assumed to be completely random, but in medical scenarios, the reality is that these patterns occur in bursts due to sensors that are off for some time or data collected in a misaligned uneven fashion, among other causes. This paper proposes to model medical data records with heterogeneous data types and bursty missing data using sequential variational autoencoders (VAEs). In particular, we propose a new methodology, the Shi-VAE, which extends the capabilities of VAEs to sequential streams of data with missing observations. We compare our model against state-of-the-art solutions in an intensive care unit database (ICU) and a dataset of passive human monitoring. Furthermore, we find that standard error metrics such as RMSE are not conclusive enough to assess temporal models and include in our analysis the cross-correlation between the ground truth and the imputed signal.
We show that Shi-VAE achieves the best performance in terms of using both metrics, with lower computational complexity than the GP-VAE model, which is the state-of-the-art method for medical records. 
\end{abstract}


\IEEEpeerreviewmaketitle

\section{Introduction}
\label{sec:introduction}
Since machine learning emerged, all the primary attention focused on working with homogeneous data sets, where too few artifacts such as outliers or missing data barely appear. But real-world data sets are quite different. Data is usually organized in databases containing incomplete, noisy, and more critical, heterogeneous information sources. These scenarios are quite common in medical applications. For instance, Electronic Health Records (EHR) may contain information from monitoring sensors, different physicians' diagnoses, or visits to the hospital. A heterogeneous medical footprint hence defines each patient. This kind of information will exhibit missing data due to sensors' failures or due to temporal gaps between each visit to the hospital, to name a few. 

\begin{figure}[t]
	\centering
	\includegraphics[width=0.3\textwidth]{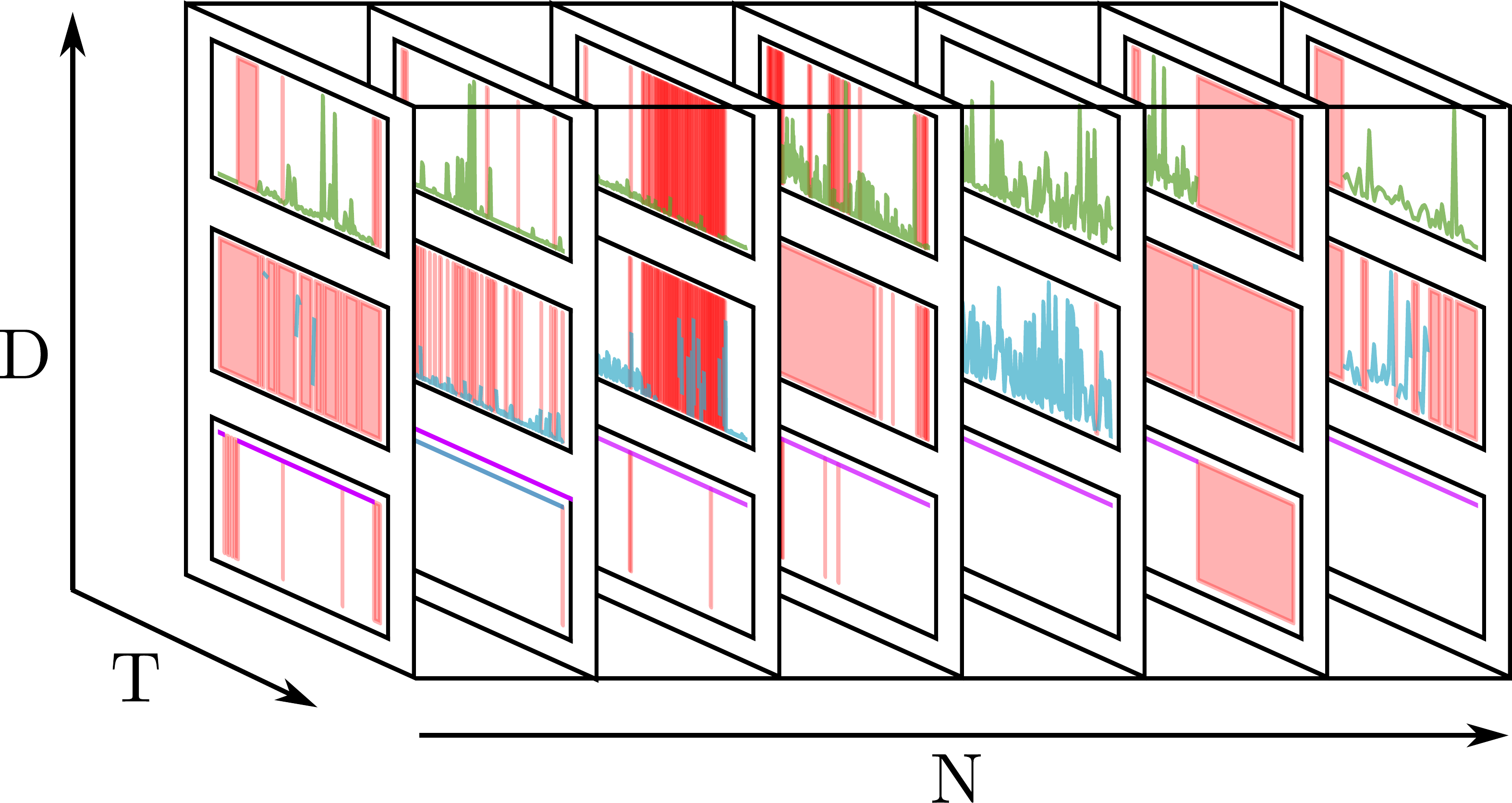}
	\caption{Example of heterogeneous streams of data with missing values from the medical data set. Red vertical lines correspond to missing values. Each row corresponds to a different type of data: the first two correspond to positive real-valued data and the third to binary data. $D$ refers to dimensionality of the dataset, $T$ to the temporal dimension and $N$ to the number of samples.}
	\label{fig:eb2slots}
\end{figure}

In the literature, the common assumption is that the lost information from a data set is Missing Completely at Random (MCAR). However, the most usual scenario is that missing data follows some kind of pattern. For example, in human monitoring applications the sensors tracking different sources might disconnect for some amount of time, not intermittently, generating \textit{bursts of missing data}. For medical data sets missing patterns can appear simultaneously across different attributes as it is shown in Figure \ref{fig:eb2slots}.

% ================= Basics
The recent literature on machine learning (ML) approaches to handle noise and missing data in medical records is dominated by deep learning methods.  In this regard, recurrent neural networks (RNN) stand as one of the most popular approaches.  In \cite{lipton2015learning} the authors propose  Long-Short-Term Memory (LSTM) networks \cite{hochreiter1997long}, to recognize patterns in multivariate time series of clinical measurements. This work was extended in \cite{lipton2016modeling} with binary indicators of missingness as features. A different approach is proposed in \cite{che2018recurrent}, where Gated Recurrent Units (GRU) are modified to incorporate missing masks, hence modeling the time intervals between clinical appointments.  Other works like BRITS \cite{cao2018brits} also look into the bidirectional capabilities of RNNs and exploit this property to impute missing values in time series with underlying nonlinear dynamics.

Although the above RNN-based methods show impressive results dealing with time series forecasting, they do not benefit from the flexibility and the underlying data correlations inferred by probabilistic deep generative models (DGMs). DGMs capture inner correlations that can be present in high-dimensional data employing a low-dimensional latent space. In the framework of VAEs, the heterogeneous incomplete variational autoencoder (HI-VAE) \cite{nazabal2020handling}, the mixed VAE (VAEM) \cite{ma2020vaem}, the MIWAE \cite{mattei2019miwae}, the Partial VAE presented in \cite{ma2018eddi} or similar works \cite{collier2020vaes} \cite{giaa082} propose efficient methods to jointly model different data types and missing data in a single DGM. Among  DGMs able to deal with sequential data, GP-VAE \cite{fortuin2020gp} stands out. GP-VAE implements a latent probabilistic model in which a Gaussian process captures the correlation of the low-dimensional latent variable along time,  and this GP relies on a VAE to implement the observation model. However, GP-VAE cannot deal with heterogeneous observations. Finally, DGM-like solutions to deal with tabular or sequential based on generative adversarial networks (GANs), such as GAIN in \cite{yoon2018gain}, the gated recurrent GAN in \cite{NIPS2018_7432}, MisGAN in \cite{li2018learning} and VIGAN \cite{shang2017vigan} do not show to outperform the imputation ability of other VAE-based methods and are harder to train due to the min-max underlying optimization problem.  

In this paper, we consider modeling sequential heterogeneous data when missing data comes in bursts, a scenario in which none of the previous DGMs have been tested to date. On the one hand, we show that when errors come in bursts, standard error metrics such as normalized mean-squared error (NRMSE) do not reflect well the imputation accuracy, and we study the correlation between the ground-truth signal and the imputed one.  In this setup, we demonstrate that GP-VAE struggles to deal with long-missing data bursts since the underlying GP correlation quickly decays, driving the GP posterior to a non-informative mean and large variance. 

To better deal with bursty missing patterns, we propose the sequential heterogeneous incomplete VAE (Shi-VAE). This model generalizes the HI-VAE model in \cite{nazabal2020handling}, including a latent temporal structure driven by LSTMs following a similar idea as in \cite{vrnn}. The extended memory properties of these networks provide a more robust ability to cope with missing bursts, efficiently capturing into the low-dimensional latent projection the correlation to past observations.  Besides, Shi-VAE comes with efficient training methods based on amortized variational inference that can handle massive data sets.  As a representative example of a medical database, we demonstrate the superior ability of Shi-VAE to deal with complex time-series using two real data sets. First, we consider the data set from the 2012 Physionet Challenge \cite{physionet} which contains measurements of 35 electrophysiological signals for 12,000 patients monitored during 48 on the intensive care unit (ICU). Second, we consider a data set of human passive monitoring coming from mobile devices. It contains heterogeneous attributes (distance travelled, mobile phone usage, quality of sleep, etc.) and a challenging presence of bursty missing data. The Shi-VAE code to reproduce our experiments can be found in \url{https://github.com/dbarrejon/Shi-VAE}. Overall, we claim the following contributions:
\begin{itemize}
    \item We propose Shi-VAE as a robust generative model to handle heterogeneous time series corrupted with missing data.
    \item We demonstrate that NRMSE is a partial metric when it comes to compare imputation models in the presence of missing data in bursts.
    \item We propose to use a temporal correlation metric to compare the different models. This metric is more sensitive to detect over-smooth solutions.
\end{itemize}

We organize the paper as follows. Firstly, Section \ref{sec:human_monitoring} introduces the problem statement we want to tackle. Section \ref{sec:model} presents Shi-VAE. In Section \ref{sec:results} we present the two data sets we have used to validate our model and the results we have found. Section \ref{sec:discussion} presents our final remarks.

\section{A human monitoring database}
\label{sec:human_monitoring}

Through patients' mobile phones and other wearable devices, continuous sensor data can be collected in a non-invasive manner, providing valuable information about everyday activity patterns. The possibility of inferring emotional states by analyzing smartphone usage data \cite{moodscope}, \cite{mytraces}, GPS traces of movement \cite{canzian2015trajectories}, social media data \cite{de2013predicting}, and even sound recordings \cite{lu2012stresssense} has become a growing research focus over the past decade.

One of the databases that we use in this paper was collected using the mobile application eB2 MindCare \cite{eb2} in a collaboration we carried out with two public mental health hospitals in Madrid (Hospital Universitario Fundación Jiménez Díaz and Hospital Universitario Rey Juan Carlos). This study was approved by the Fundación Jiménez Díaz Research Ethics Committee (Study code: LSRG-1-005 16). We periodically capture passive monitoring information from $N=170$ psychiatric patients using eB2 MindCare, thus registering different signals for every user. In particular, we are working with daily summary representations of every variable.  The seven attributes we work with are listed in Table \ref{tab:eb2_data}, along with the fraction of missing values across all patients.  

\begin{figure*}[t]
	\centering
	\includegraphics[width=1.0\textwidth]{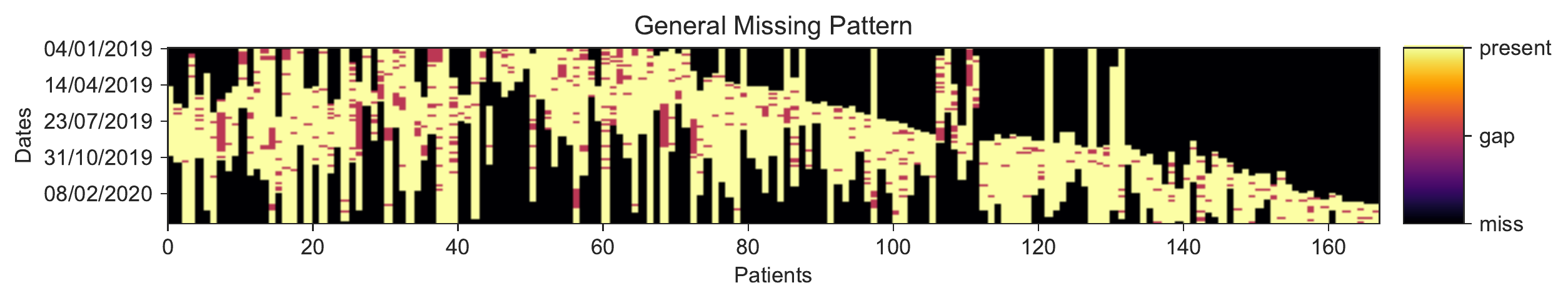}
	\caption{Overall view of the human monitoring database. Each patient has a given sequence length. Black means no record of that patient, magenta means a complete missing day and yellow that at least there is one variable present at that day.}
	\label{fig:gaps}
\end{figure*}

Regarding the positive variables, distance, steps total, and vehicle are related to the patient's mobility. App usage is a positive variable that measures the total amount of active time the user has been using the phone, with social applications, phone calls, etc. Sleep is a positive variable that counts the total time a person has slept during a day. Regarding binary variables, sport explains whether the person has done any sport $x_t=1$ or not $x_t=0 $ during the day and steps home states whether the person was at home $x_t=1$ or not $x_t=0$ at that particular day. 

\begin{table}[h]
	\centering
	\begin{tabular}{@{}llc@{}}
		\toprule
		Variable      & Type & Missing Percentage [$\%$] \\ \midrule
		Distance &   Positive &  42    \\
		Steps Home &  Binary & 66 \\
		Steps Total & Positive & 22 \\
		App Usage  & Positive & 38\\
		Sport & Binary & 62 \\
		Sleep & Positive & 31 \\
		Vehicle & Positive & 44 \\
		%Valence & Categorical & 72 \\
		\bottomrule
	\end{tabular}
	\caption{Human Monitoring data set.}
	\label{tab:eb2_data}
\end{table}

Finally, we remark that, although the number $D$ of attributes is the same for every patient ($D=7$), the signal length $T$ per patient is very diverse. The average sequence length is 233. Figure \ref{fig:gaps} illustrates the whole population and the missing pattern. From the Figure \ref{fig:gaps} we can observe that almost any day comes with missing values, and hence we can expect long bursts of missing attributes. 

In this paper, we demonstrate the superior ability of the proposed Shi-VAE to capture the non-trivial correlations among the database attributes and accurately impute missing values.

\section{Proposed Model}
\label{sec:model}

We first introduce a general notation of the problem and then present the Shi-VAE model.

\subsection{Notation}
We define our data set as $\dataset= \{\X^1, \dots, \X^N\}$, where $N$ corresponds to the total number of samples in the data set.  Each sample $\X^n \in \realset^{T^n \times d}$ has $T^n$  observations $\x_t = [x_{t1}, \dots, x_{td}]^\top \in \realset^d$, where $d$ refers to the dimension or attribute. From now on, we use $\X^n = \X$ in order to relax notation. We consider heterogeneous attributes:

\begin{itemize}
	\item \textbf{Continuous Variables}: 
	\begin{enumerate}
		\item \textbf{Real-valued data}: Data taking real values, \emph{i.e.}, $x_{td} \in \realset$.
		\item \textbf{Positive-valued data}: Data taking only positive values, \emph{i.e.}, $x_{td} \in \realset^+$.
	\end{enumerate}
	\item \textbf{Discrete Variables}:
	\begin{enumerate}
		\item \textbf{Binary Data}: Data can only be either $1$ or $0$, \emph{i.e.}, $x_{td} \in [0,1]$.
		\item \textbf{Categorical data}: Data taking values in a finite unordered set, \emph{i.e.}, $x_{td} \in \{-1,0,1\}, \, \text{or} \, x_{td} \in \{\text{'negative},\text{'neutral}, \text{'positive'}\}$.
	\end{enumerate}
\end{itemize}

Furthermore, we assume that any $\x_t$ can have both observed values and missing values. Let us define $\mathcal{O}_t$ as the index set for the observed attributes at time $t$ and $\mathcal{M}_t$ as the missing index at the same time. Hence $\mathcal{O}_t \cap \mathcal{M}_t = \emptyset$. With this notation, we can split this sentence into a vector containing observed attributes $\x_t^o$,  and a complementing vector containing missing attributes $\x_t^m$.

\subsection{The sequential heterogeneous incomplete VAE (Shi-VAE) }
This section presents the Shi-VAE probabilistic generative model, which extends the capabilities of a standard VAE to sequential heterogeneous data streams and handles missing data. In Shi-VAE, the temporal dependencies and shared correlations among attributes are captured by a latent hierarchy of low-dimensional latent variables: a continuous latent variable $\z_t \in \realset ^K$, which follows a Mixture of Gaussian's (MoG) Prior distribution \cite{dilokthanakul2016deep}, and a discrete latent variable $\s_t$ that represents the component of the MoG\footnote{Another option for the prior would be to use a mixture of posteriors as prior also known as VampPrior \cite{tomczak2018vae}. However, to us it is more reasonable to use a prior that is not dependent on the posteriors distributions, due to the implicit dependencies present in the model.}. We model the dependence between these two latent variables and the temporal data as follows:
\begin{equation}
p(\X, \Z, \mathbf{S}) = \prod_{t=1}^T p_{\theta_x}(\x_t | \z_{\leq t}, \s_t) p_{\theta_z}(\z_t|\z_{<t}, \s_{t}) p_{\theta_s}(\s_t),
\label{eq:hissvae_joint}
\end{equation}
where $\Z = \z_{\leq T}$ and $\mathbf{S} = \s_{\leq T}$. The joint probability density function is parameterized by $\theta = \{\theta_x, \theta_z, \theta_s\} $. From now on, we omit this dependency to further relax notation. Following \cite{nazabal2020handling}, we assume that given the latent variable $\z_t$ encodes all the correlation among attributes and hence they are all conditionally independent
\begin{align}
p(\x_t | \z_{\leq t}, \s_t) & =  \prod_{d \in \mathcal{O}_t}p(x_{td} | \z_{\leq t}, \s_t) \prod_{d \in  \mathcal{\mathcal{M}}_t} p(x_{td} | \z_{\leq t}, \s_t).
%& =  p(\x_t^o | \z_{\leq t}, \s_t) p(\x_t^m | \z_{\leq t}, \s_t). 
\label{eq:factorization}
\end{align}
The actual expression for each of the likelihood factors $p(x_{td} | \z_{\leq t}, \s_t)$ depends on the data-type of every attribute, as we develop in the next sub-section.

\begin{figure*}[t]
	\centering
	\includegraphics[scale=0.16]{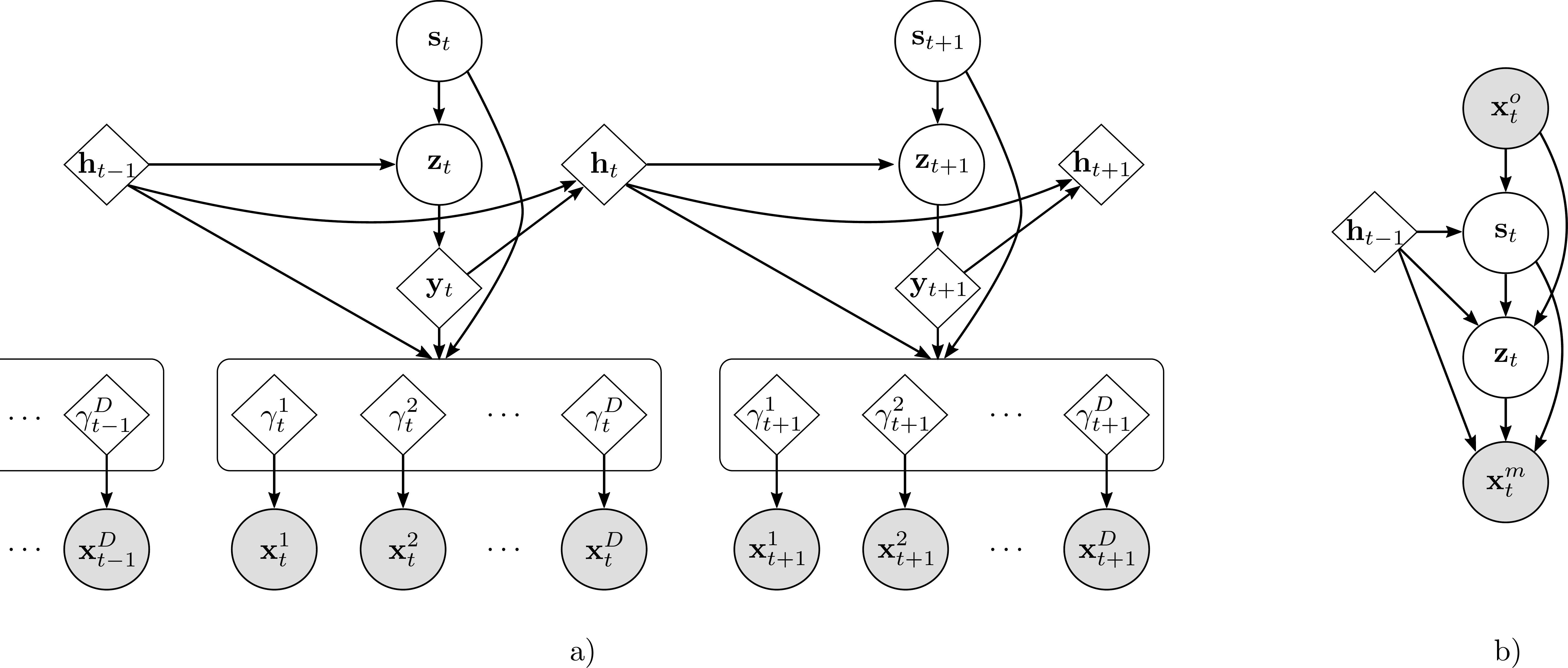}
	\caption{On a), Shi-VAE generative model. On b), Shi-VAE inference model.}
	\label{fig:tgmissvae}
\end{figure*}

The temporal dependency is encoded into the term $p_{\theta_z}(\z_t|\z_{<t}, \s_{t})$, which implements a RNN-based model to capture the temporal data correlation along time:
\begin{align}
p(\z_t | \z_{< t}, \s_t) & = \mathcal{N}(\z_t | \boldsymbol{\mu}_{0,t}, \boldsymbol{\Sigma}_{0,t}),
\label{eq:prior_z}
\end{align}
where  $\boldsymbol{\mu}_{0,t}$ and $\boldsymbol{\Sigma}_{0,t}$ define the parameters of the conditional prior distribution, and they are obtained as the output of a deep neural network (DNN) $\phiPrior(\cdot)$ that extracts  features from the past hidden state $\h_{t-1}$ and the current discrete state $\s_t$:
\begin{align}
[\boldsymbol{\mu}_{0,t}, \boldsymbol{\Sigma}_{0,t}] = \phiPrior(\h_{t-1}, \s_t), 
\end{align}
where $\boldsymbol{\Sigma}_{0,t}$ is considered a diagonal matrix. The hidden state $\h_{t-1}$ encodes the information of the process $\z$ up to time $t-1$, and it is updated along time using an LSTM with the following state update recurrence 
\begin{equation}
\h_{t-1} = f_{\boldsymbol{\tau}}(\y_{t-1}, \h_{t-2}),
\label{eq:h_t}
\end{equation}
where $\y_{t-1}=\varphi_{\omega}^\z(\z_{t-1})$ is the output of a DNN with input $\z_{t-1}$. We choose to work with LSTM \cite{hochreiter1997long} due to the ability to better cope with long sequences, but any other RNN architectures such as GRU \cite{cho2014learning} could be used. Besides, in order to prevent the exploding gradient problem that can arise in RNNs, we clip the gradients to 0.5. 

Finally, for the discrete latent variable $\s_t$ we assume an informative time-independent prior:
\begin{equation}
p(\s_t) = \text{Categorical}(\s_t | \boldsymbol{\pi}),
\label{eq:prior_s}
\end{equation}
where $\pi_k = 1 / L$, where $L$ is the number of components in the mixture.

\subsection{Heterogeneous Decoder}
\label{sec:heter_decoder}
We propose to use a factorized decoder that can handle different data-types for each attribute. A DNN is used to provide the likelihood parameters, e.g. mean and variance of a Gaussian distribution, given $\h_{t-1},\s_t$, and $\y_{t}$. We denote the likelihood parameters for the $d$-th attribute at time $t$  as $\gamma_{t}^d=\phiDec(\h_{t-1},\s_t,\y_t)$, where $\phiDec$ is the \emph{decoder} DNN, as it translates latent information into the observed variable space. Hence, the general likelihood expression is:
\begin{align}
    p(x_{td} | \z_{\leq t}, \s_t) = p(x_{td} | \gamma_{t}^d)
\label{eq:decoder}
\end{align}
We consider the following data-types and associated likelihood forms:

\begin{enumerate}
	\item \textbf{Real-valued data}: We assume a Gaussian likelihood distribution, \emph{i.e.},
	\begin{align}
	p(x_{td} | \gamma_{t}^d) =  \mathcal{N}( & \mu_{x,t}^d, \sigma_{x,t}^{2,d}), \nonumber\\
	 \text{where} \,  [& \mu_{x,t}^d, \sigma_{x,t}^{2,d}] = \phiDec (\y_t, \s_t, \h_{t-1})
	\end{align}

	\item \textbf{Positive real-valued data}: We assume a log-Gaussian likelihood distribution, \emph{i.e.},
	\begin{align}
	p(x_{td} | \gamma_{t}^d) =  \log \mathcal{N}(& \mu_{x,t}^d, \sigma_{x,t}^{2,d}), \\
	\text{where} \,  [& \mu_{x,t}^d, \sigma_{x,t}^{2,d}] = \phiDec (\y_t, \s_t, \h_{t-1})\nonumber
	\end{align}

	\item \textbf{Binomial data}: We assume a Bernoulli likelihood distribution, \emph{i.e.},
	\begin{align}
	p(x_{td} | \gamma_{t}^d) =  Be(& p_{x,t}^d), \\
	\text{where} \, & p_{x,t}^d = \sigma\big(\phiDec (\y_t, \s_t, \h_{t-1})\big), \nonumber
	\end{align}
	and $p_{x,t}^d$  is the probability parameter of the Bernoulli distribution and $\sigma$ is the sigmoid function.

	\item \textbf{Categorical data}: We assume a multinomial likelihood distribution where the parameters of the likelihood are the $C$-dimensional output of a DNN with a log-softmax output
	\begin{equation}
	\log p(x_{td} = c | \gamma_{t}^d) =  \phiDec (\y_t, \s_t, \h_{t-1})|_c
	\end{equation}	
	for $c=1,\ldots, C$.

\end{enumerate}
The left part of Figure \ref{fig:tgmissvae} illustrates the generative model defined by Equations \eqref{eq:hissvae_joint}-\eqref{eq:decoder}. From this figure we can see the motivation of having a shared latent space on $\z$ and $\s$ but an independenent heterogeneous decoder where each likelihood for $x_t^d$ is parameterized by $\gamma_t^d$.

\subsection{Model training with Variational Inference}

Variational training \cite{kingma2013auto} involves optimizing a parameterized family of distributions $q_{\eta}(\cdot)$ that approximate the latent posterior distribution given the observed data. This optimization is carried out  by maximizing the well-known evidence lower bound (ELBO).

The variational distribution for our model is defined as $q_{\phi}(\x_{\leq T}^m, \z_{\leq T}, \s_{\leq T} | \x_{\leq T}^o)$ and it only depends on the observed attributes. Firstly, we need to define the variational distribution over the latent variable $\z_t$
\begin{align}
q_{\phi_z}(\z_t | \z_{<t}, \s_t, \x_t^o) & = \mathcal{N}(\boldsymbol{\mu}_{z,t}, \boldsymbol{\Sigma}_{z,t}), \label{eq:var_z}
\\
&  \text{where} \, [\boldsymbol{\mu}_{z,t}, \boldsymbol{\Sigma}_{z,t}] = \phiEnc (\phiX(\tilde{\x}_t), \h_{t-1}, \s_t), \nonumber 
\end{align}
where $\tilde{\x}_{t}$ denotes a $D$-dimensional vector where the missing dimensions have been replaced by zeros following the zero filling approach as described in \cite{nazabal2020handling}, $\boldsymbol{\mu}_{z,t}$ and  $\boldsymbol{\Sigma}_{z,t}$ are the parameters of the variational distribution and $\phiX$ and $\phiEnc$ are neural networks. $\boldsymbol{\Sigma}_{z,t}$ is a diagonal matrix. The variational distribution for the discrete latent space $\s_t$ is defined as 
\begin{equation}
q_{\phi_s}(\s_t | \x_t^o, \z_{<t}) = \text{Categorical}(\boldsymbol{\pi}(\phiXs(\tilde{\x}_t, \h_{t-1}))),
\label{eq:var_s}
\end{equation}
where the probability for each category is given by the output of the DNN $\phiXs(\cdot)$ followed by a log soft-max function. The variational distribution will then be composed of the variational distribution from Equation \eqref{eq:var_z}, the variational distribution from Equation \eqref{eq:var_s} and $p(\x_t^m | \z_{\leq t}, \s_t)$, \emph{i.e.}
\begin{align}
q_{\phi}(\x_{\leq T}^m, \z_{\leq T}, \s_{\leq T} | \x_{\leq T}^o) = \prod\nolimits_{t=1}^{T} 
& q_{\phi_z}(\z_t | \z_{<t}, \s_t, \x_t^o) \\
& q_{\phi_s}(\s_t| \x_t^o, \z_{<t}) \nonumber\\
& p(\x_t^m | \z_{\leq t}, \s_t).\nonumber
\label{eq:inference2} 
\end{align}
The inference model is shown at the right part of Figure \ref{fig:tgmissvae}. By expanding the following expression
\begin{align}
\log p(\X^o) \geq \int q(\X^o,\X^m,\Z,\mathbf{S}) \log \frac{p(\X, \Z, \mathbf{S})}{q(\X^o,\X^m,\Z,\mathbf{S})} d\Z d\mathbf{S} d\X^m,
\end{align}
\begin{table*}[ht]
\begin{align}
\log p(\X^o) \geq \sum_{t = 1}^{T} \bigg[
\underbrace{\underset{q(\z_t | \x_t^o, \z_{<t}, \s_{t})}{\underset{q(\s_{t} | \x_{t}^o, \z_{<t},)}{\mathbb{E}}} \big[  \log p(\x_t^o | \z_{\leq t}, \s_{t})\big]}_{\text{Reconstruction}} & -  \underbrace{\underset{}{\underset{q(\s_{t} | \x_{t}^o, \z_{<t})}{\mathbb{E}}} \big[\beta \KL\big(q(\z_t | \z_{<t}, \x_t^o, \s_{t}) || p(\z_{t} | \z_{<t}, \s_{t}))\big] - \beta\KL(q(\s_t | \x_t^o, \z_{<t},)|| p(\s_{t})\big)}_{\text{Regularization}}\bigg] \label{eq:ELBO}
\end{align}
\end{table*}
we obtain the ELBO objective training function defined in Equation \eqref{eq:ELBO}.
The first term inside the sum in Equation \eqref{eq:ELBO} is the average reconstruction log-likelihood (e.g. how well we explain the observed data given the latent space induced by the approximated posterior), while the other two Kullback-Leibler (KL) divergence terms act like regularizers that penalize for posteriors far from the prior latent distributions. Although the expectation over $q(\s_t | \x_t^o)$ can be computed analytically, since $\s_t$ is a discrete variable, due to the temporal dependencies encoded on the hidden state of the RNN $\h_t$ we approximate such expectations at low complexity by sampling
from $q(\s_t | \x_t^o)$ using the Gumbel-softmax trick \cite{jang2016categorical}. Finally, in Equation \eqref{eq:ELBO} $\beta$ is a regularization parameter that we gradually increase during training, in a way the KL terms  do not dominate over the reconstruction term during the earlier stages of training. Upon training, data is normalized. Standard-scaling is used for real attributes, and also to the logarithm of positive attributes. Categorical data is one-hot encoded.

\subsection{The GP-VAE probabilistic model}
\label{sec:GP-VAE}

As discussed in the introduction, GP-VAE \cite{fortuin2020gp} stands out as the state-of-the-art VAE to handle temporal series. Before addressing the experimental section, it is relevant to compare at this point the GP-VAE probabilistic model with respect to Shi-VAE. In GP-VAE, the latent temporal variable $\mathbf{z}_t$ is modeled with a  Gaussian Process (GP) \cite{rasmussen2003gaussian}, i.e., $\z_t \sim \mathcal{G} \mathcal{P}\left(m_{z}(\cdot), k_{z}(\cdot, \cdot)\right)$. The GP prior on the latent space is flexible and robust but it comes at the cost of inverting the kernel matrix, which has a time complexity of $\mathcal{O}(T^3)$. In contrast, the RNN-based correlation model in \eqref{eq:prior_z} comes with a computational cost that grows linearly in $T$. Moreover, designing a kernel function for GP-VAE that accurately captures correlations in feature space and also in the temporal dimension is challenging. 

As in Shi-VAE, in GP-VAE given $\z_t$ all the attributes are conditionally independent. Indeed, the GP-VAE and its inference machinery \cite{fortuin2020gp} does not consider heterogeneous observations, and all observations are modelled with real-valued Gaussian distributions.

\section{Experimental Results}
\label{sec:results}

In this section we test the ability of Shi-VAE to exploit hidden correlations between attributes and infer trustworthy reconstructions in the presence of missing bursts. The following models are tested against Shi-VAE in the different experiments:
\begin{itemize}
	\item \textbf{Mean}: We replace the missing values with the mean corresponding to the subsampled signal.
	\item \textbf{Last Obs Carried Forward (LOCF)}: We impute using the last observed value for a given attribute. 
	\item \textbf{KNN}: We use \textit{k}-nearest neighbor with normalized Euclidean distance to find similar samples, and then impute with a weighted average of the neighbors. 
	\item \textbf{Matrix Factorization (MF)}: We subsample and factorize the data into two low-rank matrices and impute the missing entries with matrix completion \cite{friedman2001elements}.
	\item \textbf{MICE}: We use Multiple Imputation by Chained Equations (MICE), a very common method for missing value imputation which imputes those missing values from multiple imputations with chained equations \cite{white2011multiple}.
	\item \textbf{GP-VAE}: The GP-VAE described in Section \ref{sec:GP-VAE}.
\end{itemize}

We remark that both MF and MICE are ``genie-aided" in the sense that they observe future values of the signal with-in a window to impute the results. The rest of the algorithms perform missing data imputation in an on-line fashion. Both GP-VAE and Shi-VAE reconstruct missing values by projecting the observed sequence to the latent space and then reconstruct the missing values using the generative model. The following python packages were used in order to implement the following methods: \textit{fancyimpute} for Mean, KNN and MF; \textit{autoimpute} for LOCF and \textit{scikit-learn} for MICE  \footnote{\href{https://pypi.org/project/autoimpute/}{autoimpute(0.12.1)\cite{autoimpute}}, \href{https://github.com/iskandr/fancyimpute}{fancyimpute(0.5.5)\cite{fancyimpute}}, \href{https://scikit-learn.org/stable/modules/generated/sklearn.impute.IterativeImputer.html}{mice(0.23.2)\cite{mice}}}. 

\begin{figure}[h]
	\centering
	\includegraphics[width=0.45\textwidth]{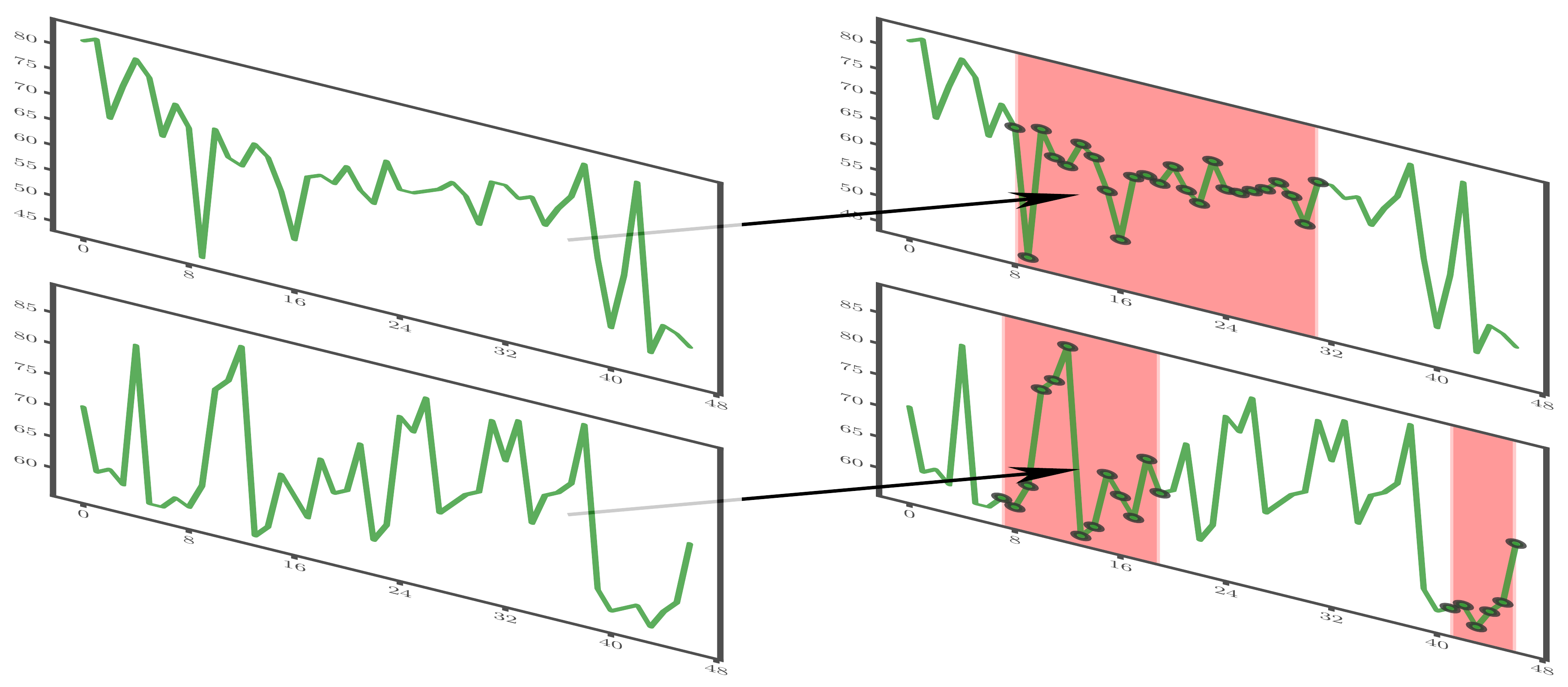}
	\caption{Generation of missing artificial bursts for different sequences. The red masks and the corresponding missing entries (black markers) indicate the bursts of missing data.} 
	\label{fig:correlation}
\end{figure}

We show results for three data sets. First, a synthetic data set generated by a heterogeneous HMM (Hidden Markov Model) with large hidden space, the human monitoring database described in Section \ref{sec:human_monitoring}, and the well known medical data set Physionet \cite{physionet}. While in the first database, the generated data set does not contain any missing data, note that both Physionet and the human monitoring database have quite a lot of missing observations. We evaluate performance over artificial missing data that we further incorporate into the data streams in all cases. We introduce missing sequences of random length for every variable to emulate missing bursts. A visual example can be seen in Figure \ref{fig:correlation}. Each burst is generated sampling a random length from a uniform distribution $\mathcal{U}(3,10)$ and placing the burst in a random position given by an observed value. For every case (database and \% introduced missing data), we create 10 random masks with a different missing pattern each, that we use to compute average errors and standard deviations around them. All masks implemented in the experiments are accessible in the code repository \url{https://github.com/dbarrejon/Shi-VAE}.

\begin{figure*}[t]
	\centering
	\includegraphics[width=0.9\textwidth]{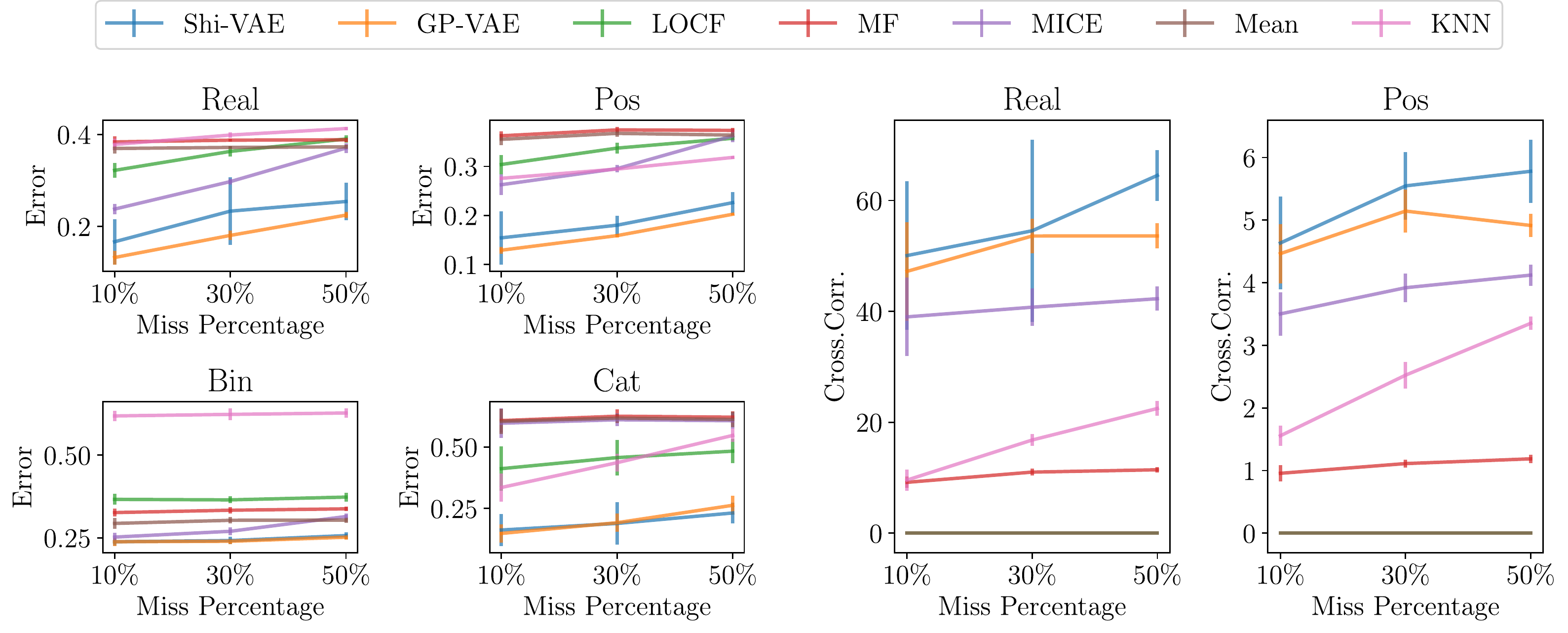}
	\caption{Results for the synthetic data set. On the left part we show the imputation error for each variable and on the right part the cross correlation for the continuous variables at different missing rates.}
	\label{fig:results_hmm}
\end{figure*}

We used the default setups for all the baselines model except for the GP-VAE, where we set the latent dimension to $2$ in the synthetic data set, to $35$ for Physionet and to $5$ for the other two databases, since these values provided optimized results after cross-validation. The cross-validated parameter configuration for the Shi-VAE is described in Table \ref{tab:params}.

\subsection{Evaluation Metrics}
We use two different types of metrics to compare our models: standard error metrics and cross-correlation metrics between the ground-truth sequence and the reconstructed one. Before presenting the evaluation metrics, we will introduce some basic notation. Let us define $\X_d$ as a $N\times T$ matrix where we compact the $d$-th attribute across all data points and time. This is the matrix before introducing the artificial missing bursts. The imputed matrix for such attribute is defined as $\hat{\X}_d$ (equal to $\X_d$ for non-missing entries). Therefore, $x_{td}^n$ is the entry at time $t$ and data point $n$ of $\X_d$. $N_d$ is the number of missing entries in $\X_d$.

\subsubsection{Error metrics}: We use a different type of error depending on the type of data:
\begin{itemize}
	\item \textbf{Continuous data}, \emph{i.e.} real and positive: we consider the normalized root mean squared error (NRMSE) evaluated only at missing entries
	\begin{equation}
	err(d) = \frac{\sqrt{1/N_d \sum_{n} \sum_{t} (x_{td}^n - \hat{x}_{td}^n)^2}}{\text{max}(\X_d) - \text{min}(\X_d)}.
	\end{equation}

	\item \textbf{Binary data and categorical data}: we consider the classification accuracy error evaluated at the missing entries.
	\begin{equation}
	err(d) = \frac{1}{N_d} \sum_{n} \sum_{t} I(x_{td}^n \ne \hat{x}_{td}^n),
	\end{equation}
	where $I(\cdot)$ is the indicator function.

\end{itemize}
The average imputation error for all the attributes is given by $\text{Error} = 1 / D \sum_{d} err(d)$, where $D$ is the number of attributes.

\subsubsection{Cross Correlation}: On temporal data sets, evaluating the performance of a given model based on standard error metrics  might not be conclusive enough, as our experiments demonstrate. We augment our experiments by analyzing $\phi(d)$, which is defined as the sum of the cross correlation between any missing burst in $\X_d$ (a portion of a given row) and its corresponding imputation in $\hat{\X}_d$, normalized by the total number of missing entries $N_d$. To simplify notation, assume $\mathbf{w}$ and $\hat{\mathbf{w}}$ are the true and imputed values of a missing burst respectively in $\X_d$, then we accumulate in $c(\mathbf{w},\hat{\mathbf{w}})$ the maximum value of the normalized cross correlation, i.e.
\begin{align}
c(\mathbf{w},\hat{\mathbf{w}}) = \max[(\mathbf{w} - \mu_{\mathbf{w}}) \star (\hat{\mathbf{w}} - \mu_{\hat{\mathbf{w}}})],
\label{xcorr}
\end{align}
$\star$ is the cross correlation operator, and $\mu_{\mathbf{w}}$ is the average signal value during the burst.

\begin{table}
	\centering
	\begin{tabular}{@{}lccc@{}}
		\toprule
		Parameter & Synthetic & Physionet & Human Monitoring\\ \midrule
		Epochs & 100 & 100 & 100\\
		Annealing Epochs & 20 & 20 & 50\\
		Dimension $\z$ & 2 & 35 & 5\\
		Dimension $\h$ & 10 & 10 & 10\\
		$L$ & 3 & 10 & 3\\
		$T$ & 100 & 48 & - \\
		Optimizer & Adam & Adam & Adam\\
		Learning Rate & $5e-3$ & $5e-3$ & $5e-3$\\
		Activation Layers & ReLU & ReLU & ReLU\\
		Split Train/Val/Test & 800/100/100 & 4K/4K/4K & 135/15/17\\
		Batch Size & 64 & 64 & 64\\
		\bottomrule
	\end{tabular}
	\caption{Parameter configuration for the different experiments.}
	\label{tab:params}
\end{table}

Hence
\begin{align}
    \phi(d) = \frac{\sum\limits_{\mathbf{w},\hat{\mathbf{w}} \in \X_d} c(\mathbf{w},\hat{\mathbf{w}})}{N_d}
\end{align}
We also report the average correlation across all attributes, \emph{i.e.} $\text{Cross. Corr} = 1 / D \sum_{d} \phi(d)$.

\subsection{Synthetic Data set}

This data set is composed of $N=1000$ samples  of length $T=100$ from a three-state HMM model. At each time instant the HMM produces four outputs of different nature: real, positive, binary and categorical. Each state is characterized by different emission distribution for each data type.
The transition probabilities have been forced to be smooth, so that really abrupt changes are not likely to happen. Over the clean database, we generate missing masks with overall missing rates of $10\%$, $30\%$ and $50\%$. For all the baselines, including the GP-VAE, we work with subsampled slots of length $T=50$ of every individual signal. For the Shi-VAE, we consider the whole signal.

In Figure \ref{fig:results_hmm} we display both reconstruction errors per attribute at different missing rates (a) and cross correlation for the real and positive attributes (b). In terms of reconstruction error,  GP-VAE obtains the best results for the continuous variables by a small margin compared to Shi-VAE. This is due to the GP-VAE assuming a fully Gaussian distribution. However, for the binary and categorical view their performance is the same. Their distance with respect to the other baselines is remarkable. On the other hand, in terms of cross correlation, observe in Figure \ref{fig:results_hmm} (b) that Shi-VAE is able to reconstruct signals that are more correlated to the true distribution of the data. This raises an important question on how temporal models that are explicitly designed to impute missing values should be analyzed, whether it is more important to just focus on standard error metrics, or metrics considering temporal dependencies should be used when assessing the validity of temporal models. 

\subsection{Physionet}

\begin{table}[t]
	\centering
	\begin{tabular}{lccc}
		\toprule
		Model &  Avg. Error  & Cross. Corr\\
		\midrule
		Shi-VAE & \textbf{0.064} $\boldsymbol{\pm}$ \textbf{0.003} &\textbf{38.061} $\boldsymbol{\pm}$ \textbf{5.000}\\
		GP-VAE & \textbf{0.060} $\boldsymbol{\pm}$ \textbf{0.002} & 31.414 $\pm$ 1.016 \\
		\bottomrule
	\end{tabular}
\caption{Physionet database results on the test set. For average error, lower is better. For cross correlation, larger is better. } \label{Tablephysionet}
\end{table}

In this section, we compare both GP-VAE and Shi-VAE over the Physionet database \cite{physionet}. The data set contains 12,000 patients which were monitored on the intensive care unit (ICU) for 48 hours each. Each signal is sampled once an hour, hence their length is $T=48$. At each hour, there is a measurement of 35 different variables\footnote{The list and definition of the attributes can be found in \cite{physionet}.} (heart rate, blood pressure, etc.), any number of which might be missing. We further introduce artificial burst missing data up to an overall fraction of 10\%. Note that the dataset already contains a large fraction of missing values.

\begin{figure*}[t]
		\centering
		\includegraphics[width=\textwidth]{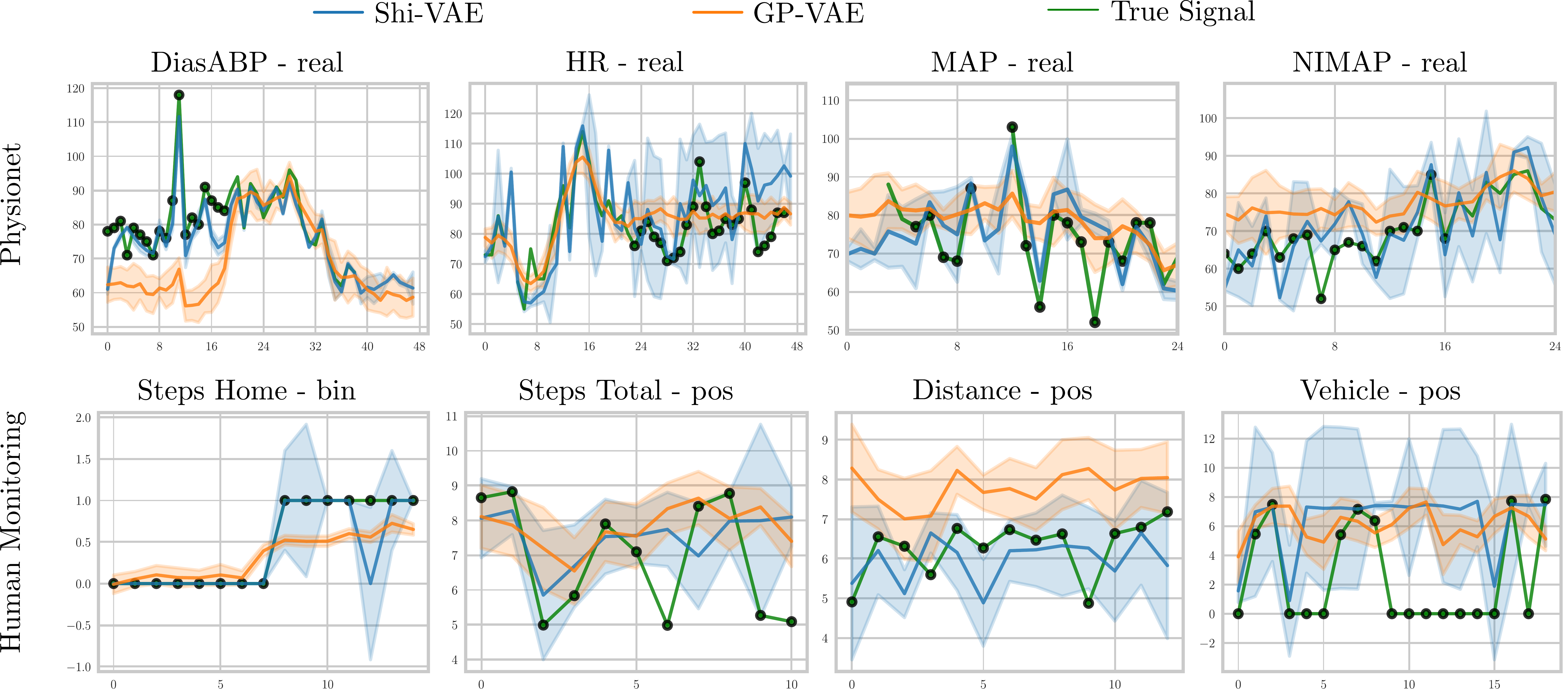}
		\caption{Shi-VAE and GP-VAE example reconstruction for different attributes over the Physionet dataset (upper row) and the human monitoring database (bottom row). Missing values are indicated by markers in the true signal.}
		\label{fig:eb2results}
\end{figure*}

In Table \ref{Tablephysionet} we report GP-VAE and Shi-VAE average reconstruction error and average cross correlation. Observe that, as in the previous case, GP-VAE slightly improves the Shi-VAE in terms of average imputation error. However, Shi-VAE achieves a larger cross-correlation with respect to the ground-truth. To illustrate why reconstruction error can be a missleading metric when it comes to missing bursts,  in the first row of Figure \ref{fig:eb2results} we display the imputation of both methods for different missing bursts located at different Physionet attributes. Missing values are indicated by markers in the true signal. Observe that, while GP-VAE tends to impute missing burst with smooth solutions, Shi-VAE imputations certainly follow the true dynamics of the signal. But this discrepancy is not reflected in the average reconstruction error.
In addition, observe that the Shi-VAE uncertainty (shaded area around the imputed signal) is informative and varies along time, allowing to identify regions of large and small uncertainty. On the other hand, the GP-VAE uncertainty does not show such a desired behaviour.

\begin{figure}[h]
		\centering
		\includegraphics[width=0.45\textwidth]{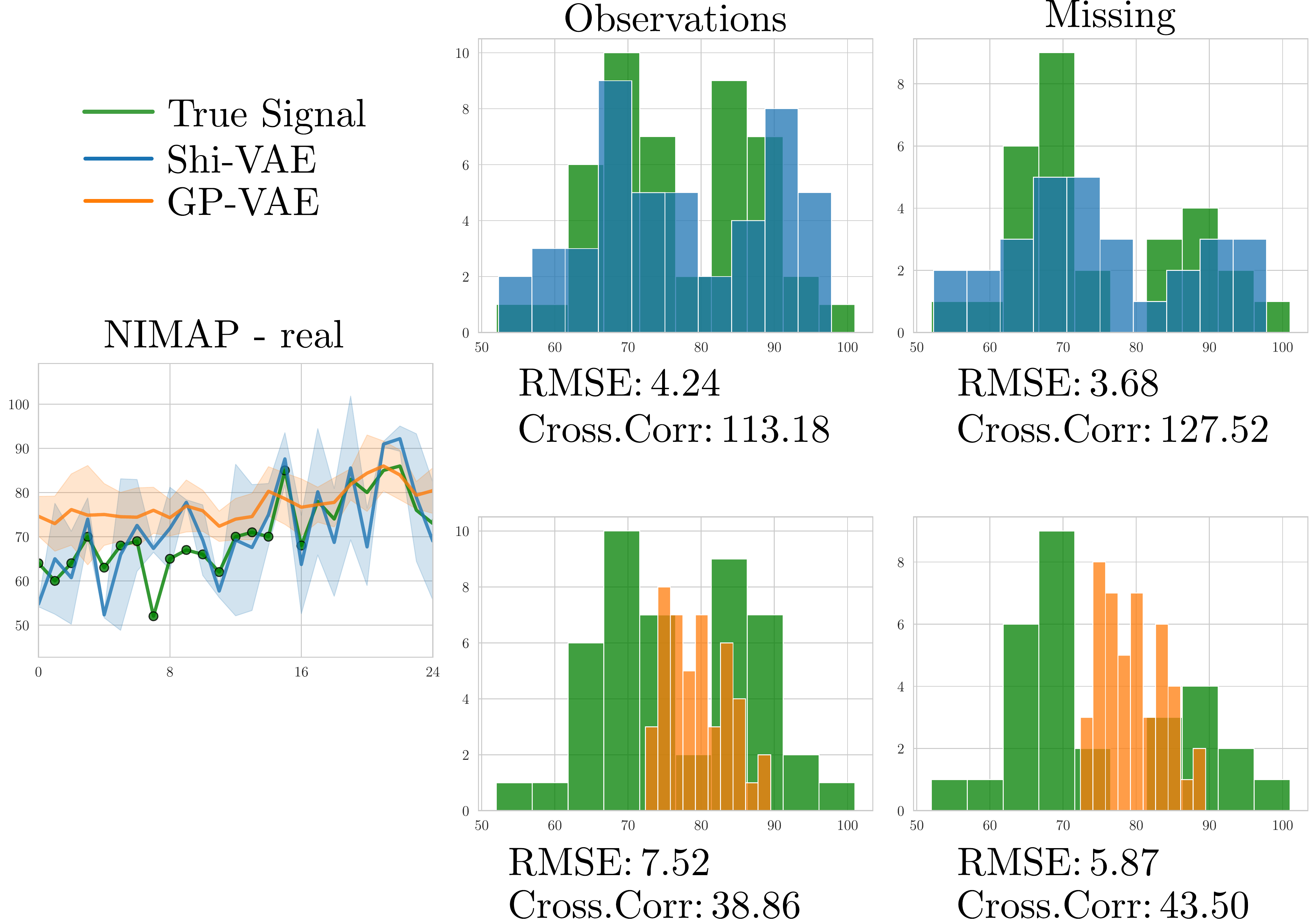}
		\caption{Comparison of Shi-VAE and GP-VAE using histograms and the evaluation metrics on missing and observed data. For RMSE (lower is better) and for cross-correlation metric (larger is better).}
		\label{fig:histogram}
\end{figure}

Similar conclusions can be drawn from the next related experiment. In Figure \ref{fig:histogram} we show one signal from the Physionet dataset, the real signal in green, the imputation from the Shi-VAE in blue and the imputation from the GP-VAE in orange. The first column on the right of each signal shows the distribution of the data with a histogram for the observed values of the signal, and the second column the distribution for the  missing values (shown with black markers on the plot). The first row corresponds to the histogram that Shi-VAE produces, obtained by sampling from the model at each point. In the second row we do the same for the GP-VAE. We use the average of 10 samples produced by the models for the results. Below each histogram we show the corresponding average RMSE between the real samples and the imputed samples for each model. Observe that, while GP-VAE struggles to fit the real distribution even in the observed values, Shi-VAE provides a reasonably better result, being able to fit the two modes of the real distribution. This issue is not clearly reflected in the RMSE metric, which is not indeed very different between both models. On the contrary, the temporal correlation metric clearly shows the superior performance of Shi-VAE.

\subsection{Human Monitoring Database}

Finally, we reproduce the experiment for the  human monitoring database described in Section \ref{sec:human_monitoring}. The average fraction of artificially introduced missing rate per attribute is $15\%$.
In this case the length of the temporal sequences for each patient is different. For Shi-VAE and GP-VAE, we pad with zeros to the right those sequences with a length smaller than the maximum sequence length in a batch.
As described in Section \ref{sec:GP-VAE}, the GP-VAE complexity badly scales with the sequence length. To run GP-VAE in reasonably time,  any sequence larger than 50 time steps is subsampled to fit this maximum length. Note that Shi-VAE does not suffer from such penalization with respect to sequence length.

In Table \ref{eb2table}, we report the error and cross correlation per attribute (seven of them, as described in Table \ref{tab:eb2_data}), and the overall average values. Observe that, systematically, Shi-VAE achieves the largest correlation per attribute. In the second row of Figure \ref{fig:eb2results} we show the imputation of both methods for different missing bursts located at different attributes. The robustness of the Shi-VAE can be observed in terms of the correlation between the imputed signal and the true one and in terms of the uncertainty along time, which tends to be larger for those points in time for which the Shi-VAE mode is far from the true value. Again, such a behavior is not provided by GP-VAE. 

\begin{table}[t]
	\centering
    \begin{tabular}{llcc}
    \toprule
    Variable & Model & Error & Cross Correlation \\
    \midrule
    Average & Shi-VAE & \textbf{0.200} $\boldsymbol{\pm}$ \textbf{0.038} & \textbf{0.369}$\boldsymbol{\pm}$ \textbf{0.140} \\
            & GP-VAE & \textbf{0.184} $\boldsymbol{\pm}$ \textbf{0.022} & 0.157 $\pm$ 0.031 \\
    \midrule            
    \midrule
    Distance & Shi-VAE & \textbf{ 0.201} $\boldsymbol{\pm}$ \textbf{0.012} & \textbf{0.783} $\boldsymbol{\pm}$ \textbf{0.249} \\
            & GP-VAE & \textbf{0.205} $\boldsymbol{\pm}$ \textbf{0.014} & 0.389 $\pm$ 0.092 \\
    \midrule
    Steps home & Shi-VAE & \textbf{0.170} $\boldsymbol{\pm}$ \textbf{0.054} & \textbf{0.010} $\boldsymbol{\pm}$ \textbf{0.009} \\
            & GP-VAE & \textbf{0.151} $\boldsymbol{\pm}$ \textbf{0.016} & \textbf{0.011} $\boldsymbol{\pm}$ \textbf{0.009}\\
    \midrule
    Steps total & Shi-VAE & \textbf{0.269} $\boldsymbol{\pm}$ \textbf{0.046} & \textbf{0.444} $\boldsymbol{\pm}$ \textbf{0.181} \\
            & GP-VAE & \textbf{0.268} $\boldsymbol{\pm}$ \textbf{0.044} & 0.205 $\pm$ 0.038 \\
    \midrule
    App usage & Shi-VAE & \textbf{0.113} $\boldsymbol{\pm}$ \textbf{0.014} & \textbf{0.088} $\boldsymbol{\pm}$ \textbf{0.045} \\
            & GP-VAE & \textbf{0.115} $\boldsymbol{\pm}$ \textbf{0.013} & \textbf{0.039} $\boldsymbol{\pm}$ \textbf{0.008} \\
    \midrule
    Sport & Shi-VAE & \textbf{0.216} $\boldsymbol{\pm}$ \textbf{0.086} & \textbf{0.013} $\boldsymbol{\pm}$ \textbf{0.005} \\
            & GP-VAE & \textbf{0.121} $\boldsymbol{\pm}$ \textbf{0.030} & \textbf{0.009} $\boldsymbol{\pm}$ \textbf{0.004} \\
    \midrule
    Sleep & Shi-VAE & \textbf{0.063} $\boldsymbol{\pm}$ \textbf{0.010} & \textbf{0.034} $\boldsymbol{\pm}$ \textbf{0.016} \\
            & GP-VAE & \textbf{0.059} $\boldsymbol{\pm}$ \textbf{0.010} & 0.013 $\pm$ 0.003 \\
    \midrule
    Vehicle & Shi-VAE & \textbf{0.372} $\boldsymbol{\pm}$ \textbf{0.043} & \textbf{1.215} $\boldsymbol{\pm}$ \textbf{0.477} \\
            & GP-VAE & \textbf{0.370} $\boldsymbol{\pm}$ \textbf{0.028} & 0.436 $\pm$ 0.064 \\
    \bottomrule
    \end{tabular}
\caption{Results for each variable for the human monitoring data set.}\label{eb2table}
\end{table}

\section{Discussion}
\label{sec:discussion}

In this work we propose Shi-VAE, a deep generative model that handles temporal and heterogeneous streams of data in the presence of missing data.
While GP-VAE badly scales with long time series, Shi-VAE handles long term dependencies by encapsulating the temporal information into the continuous latent code $\z$ by using RNN architectures. Having a hierarchical latent model with an additional discrete latent embedding $\s$ provides a more flexible understanding of the data and benefits the latter process of modeling the heterogeneous distributions. 

We have shown with a synthetic data set and two real-world medical data sets that standard error metrics are not completely informative to fully assess the performance of temporal models. We remark the importance of analyzing the temporal correlation in these type of studies by using sequences of missing data along time instead of fully random missing masks as it is normally done in similar works. In this scenario, Shi-VAE emerges as a robust solution to impute missing data bursts and perform dimensionality reduction.

% References
\bibliographystyle{IEEEtran}
\bibliography{bibliography} 

% Generated by IEEEtran.bst, version: 1.14 (2015/08/26)
\begin{thebibliography}{10}
\providecommand{\url}[1]{#1}
\csname url@samestyle\endcsname
\providecommand{\newblock}{\relax}
\providecommand{\bibinfo}[2]{#2}
\providecommand{\BIBentrySTDinterwordspacing}{\spaceskip=0pt\relax}
\providecommand{\BIBentryALTinterwordstretchfactor}{4}
\providecommand{\BIBentryALTinterwordspacing}{\spaceskip=\fontdimen2\font plus
\BIBentryALTinterwordstretchfactor\fontdimen3\font minus
  \fontdimen4\font\relax}
\providecommand{\BIBforeignlanguage}[2]{{%
\expandafter\ifx\csname l@#1\endcsname\relax
\typeout{** WARNING: IEEEtran.bst: No hyphenation pattern has been}%
\typeout{** loaded for the language `#1'. Using the pattern for}%
\typeout{** the default language instead.}%
\else
\language=\csname l@#1\endcsname
\fi
#2}}
\providecommand{\BIBdecl}{\relax}
\BIBdecl

\bibitem{lipton2015learning}
Z.~C. Lipton, D.~C. Kale, C.~Elkan, and R.~Wetzel, ``Learning to diagnose with
  {LSTM} recurrent neural networks,'' in \emph{Proc. ICLR}, 2016, pp. 1--18.

\bibitem{hochreiter1997long}
S.~Hochreiter and J.~Schmidhuber, ``Long short-term memory,'' \emph{Neural
  computation}, vol.~9, no.~8, pp. 1735--1780, 1997.

\bibitem{lipton2016modeling}
Z.~C. Lipton, D.~C. Kale, R.~Wetzel \emph{et~al.}, ``Modeling missing data in
  clinical time series with rnns,'' \emph{Machine Learning for Healthcare},
  vol.~56, 2016.

\bibitem{che2018recurrent}
Z.~Che, S.~Purushotham, K.~Cho, D.~Sontag, and Y.~Liu, ``Recurrent neural
  networks for multivariate time series with missing values,'' \emph{Scientific
  reports}, vol.~8, no.~1, pp. 1--12, 2018.

\bibitem{cao2018brits}
\BIBentryALTinterwordspacing
W.~Cao, D.~Wang, J.~Li, H.~Zhou, L.~Li, and Y.~Li, ``Brits: Bidirectional
  recurrent imputation for time series,'' in \emph{Advances in Neural
  Information Processing Systems}, S.~Bengio, H.~Wallach, H.~Larochelle,
  K.~Grauman, N.~Cesa-Bianchi, and R.~Garnett, Eds., vol.~31.\hskip 1em plus
  0.5em minus 0.4em\relax Curran Associates, Inc., 2018. [Online]. Available:
  \url{https://proceedings.neurips.cc/paper/2018/file/734e6bfcd358e25ac1db0a4241b95651-Paper.pdf}
\BIBentrySTDinterwordspacing

\bibitem{nazabal2020handling}
A.~Nazabal, P.~M. Olmos, Z.~Ghahramani, and I.~Valera, ``Handling incomplete
  heterogeneous data using {VAE}s,'' \emph{Pattern Recognition}, p. 107501,
  2020.

\bibitem{ma2020vaem}
\BIBentryALTinterwordspacing
C.~Ma, S.~Tschiatschek, R.~Turner, J.~M. Hern\'{a}ndez-Lobato, and C.~Zhang,
  ``{VAEM}: a deep generative model for heterogeneous mixed type data,'' in
  \emph{Advances in Neural Information Processing Systems}, H.~Larochelle,
  M.~Ranzato, R.~Hadsell, M.~F. Balcan, and H.~Lin, Eds., vol.~33.\hskip 1em
  plus 0.5em minus 0.4em\relax Curran Associates, Inc., 2020, pp.
  11\,237--11\,247. [Online]. Available:
  \url{https://proceedings.neurips.cc/paper/2020/file/8171ac2c5544a5cb54ac0f38bf477af4-Paper.pdf}
\BIBentrySTDinterwordspacing

\bibitem{mattei2019miwae}
P.-A. Mattei and J.~Frellsen, ``Miwae: Deep generative modelling and imputation
  of incomplete data sets,'' in \emph{International Conference on Machine
  Learning}, 2019, pp. 4413--4423.

\bibitem{ma2018eddi}
C.~Ma, S.~Tschiatschek, K.~Palla, J.~M. Hern{\'a}ndez-Lobato, S.~Nowozin, and
  C.~Zhang, ``Eddi: Efficient dynamic discovery of high-value information with
  partial {VAE},'' \emph{arXiv preprint arXiv:1809.11142}, 2018.

\bibitem{collier2020vaes}
M.~Collier, A.~Nazabal, and C.~K.I.~Williams, ``{VAE}s in the presence of
  missing data,'' in \emph{International Conference on Machine Learning 2020
  Workshop Art of Learning with Mising Values (Artemiss)}.

\bibitem{giaa082}
Y.~L. Qiu, H.~Zheng, and O.~Gevaert, ``{Genomic data imputation with
  variational auto-encoders},'' \emph{GigaScience}, vol.~9, no.~8, 08 2020.

\bibitem{fortuin2020gp}
V.~Fortuin, D.~Baranchuk, G.~R{\"a}tsch, and S.~Mandt, ``{GP-VAE}: Deep
  probabilistic time series imputation,'' in \emph{International Conference on
  Artificial Intelligence and Statistics}.\hskip 1em plus 0.5em minus
  0.4em\relax PMLR, 2020, pp. 1651--1661.

\bibitem{yoon2018gain}
J.~Yoon, J.~Jordon, and M.~van~der Schaar, ``{GAIN}: Missing data imputation
  using generative adversarial nets,'' in \emph{Proceedings of the 35th
  International Conference on Machine Learning}, ser. Proceedings of Machine
  Learning Research, vol.~80.\hskip 1em plus 0.5em minus 0.4em\relax PMLR,
  2018, pp. 5689--5698.

\bibitem{NIPS2018_7432}
\BIBentryALTinterwordspacing
Y.~Luo, X.~Cai, Y.~ZHANG, J.~Xu, and Y.~xiaojie, ``Multivariate time series
  imputation with generative adversarial networks,'' in \emph{Advances in
  Neural Information Processing Systems}, S.~Bengio, H.~Wallach, H.~Larochelle,
  K.~Grauman, N.~Cesa-Bianchi, and R.~Garnett, Eds., vol.~31.\hskip 1em plus
  0.5em minus 0.4em\relax Curran Associates, Inc., 2018. [Online]. Available:
  \url{https://proceedings.neurips.cc/paper/2018/file/96b9bff013acedfb1d140579e2fbeb63-Paper.pdf}
\BIBentrySTDinterwordspacing

\bibitem{li2018learning}
S.~C.-X. Li, B.~Jiang, and B.~Marlin, ``Misgan: Learning from incomplete data
  with generative adversarial networks,'' in \emph{International Conference on
  Learning Representations}, 2019.

\bibitem{shang2017vigan}
C.~Shang, A.~Palmer, J.~Sun, K.-S. Chen, J.~Lu, and J.~Bi, ``Vigan: Missing
  view imputation with generative adversarial networks,'' in \emph{2017 IEEE
  International Conference on Big Data (Big Data)}.\hskip 1em plus 0.5em minus
  0.4em\relax IEEE, 2017, pp. 766--775.

\bibitem{vrnn}
\BIBentryALTinterwordspacing
J.~Chung, K.~Kastner, L.~Dinh, K.~Goel, A.~C. Courville, and Y.~Bengio, ``A
  recurrent latent variable model for sequential data,'' in \emph{Advances in
  Neural Information Processing Systems}, C.~Cortes, N.~Lawrence, D.~Lee,
  M.~Sugiyama, and R.~Garnett, Eds., vol.~28.\hskip 1em plus 0.5em minus
  0.4em\relax Curran Associates, Inc., 2015. [Online]. Available:
  \url{https://proceedings.neurips.cc/paper/2015/file/b618c3210e934362ac261db280128c22-Paper.pdf}
\BIBentrySTDinterwordspacing

\bibitem{physionet}
I.~Silva, G.~Moody, D.~J. Scott, L.~A. Celi, and R.~G. Mark, ``Predicting
  in-hospital mortality of icu patients: The physionet/computing in cardiology
  challenge 2012,'' in \emph{2012 Computing in Cardiology}.\hskip 1em plus
  0.5em minus 0.4em\relax IEEE, 2012, pp. 245--248.

\bibitem{moodscope}
R.~LiKamWa, Y.~Liu, N.~D. Lane, and L.~Zhong, ``Moodscope: Building a mood
  sensor from smartphone usage patterns,'' in \emph{Proceeding of the 11th
  annual international conference on Mobile systems, applications, and
  services}, 2013, pp. 389--402.

\bibitem{mytraces}
A.~Mehrotra, F.~Tsapeli, R.~Hendley, and M.~Musolesi, ``Mytraces: Investigating
  correlation and causation between users’ emotional states and mobile phone
  interaction,'' \emph{Proceedings of the ACM on Interactive, Mobile, Wearable
  and Ubiquitous Technologies}, vol.~1, no.~3, pp. 1--21, 2017.

\bibitem{canzian2015trajectories}
L.~Canzian and M.~Musolesi, ``Trajectories of depression: unobtrusive
  monitoring of depressive states by means of smartphone mobility traces
  analysis,'' in \emph{Proceedings of the 2015 ACM international joint
  conference on pervasive and ubiquitous computing}, 2015, pp. 1293--1304.

\bibitem{de2013predicting}
M.~De~Choudhury, S.~Counts, and E.~Horvitz, ``Predicting postpartum changes in
  emotion and behavior via social media,'' in \emph{Proceedings of the SIGCHI
  conference on human factors in computing systems}, 2013, pp. 3267--3276.

\bibitem{lu2012stresssense}
H.~Lu, D.~Frauendorfer, M.~Rabbi, M.~S. Mast, G.~T. Chittaranjan, A.~T.
  Campbell, D.~Gatica-Perez, and T.~Choudhury, ``Stresssense: Detecting stress
  in unconstrained acoustic environments using smartphones,'' in
  \emph{Proceedings of the 2012 ACM conference on ubiquitous computing}, 2012,
  pp. 351--360.

\bibitem{eb2}
``Evidence base behavior (eb2),'' https://eb2.tech/.

\bibitem{dilokthanakul2016deep}
N.~Dilokthanakul, P.~A.~M. Mediano, M.~Garnelo, M.~C.~H. Lee, H.~Salimbeni,
  K.~Arulkumaran, and M.~Shanahan, ``Deep unsupervised clustering with gaussian
  mixture variational autoencoders.'' \emph{CoRR}, vol. abs/1611.02648, 2016.

\bibitem{tomczak2018vae}
J.~Tomczak and M.~Welling, ``{VAE} with a vampprior,'' in \emph{International
  Conference on Artificial Intelligence and Statistics}.\hskip 1em plus 0.5em
  minus 0.4em\relax PMLR, 2018, pp. 1214--1223.

\bibitem{cho2014learning}
\BIBentryALTinterwordspacing
K.~Cho, B.~van Merrienboer, {\c{C}}.~G{\"{u}}l{\c{c}}ehre, F.~Bougares,
  H.~Schwenk, and Y.~Bengio, ``Learning phrase representations using {RNN}
  encoder-decoder for statistical machine translation,'' \emph{CoRR}, vol.
  abs/1406.1078, 2014. [Online]. Available:
  \url{http://arxiv.org/abs/1406.1078}
\BIBentrySTDinterwordspacing

\bibitem{kingma2013auto}
D.~P. Kingma and M.~Welling, ``{Auto-Encoding Variational Bayes},'' in
  \emph{2nd International Conference on Learning Representations, {ICLR} 2014,
  Banff, AB, Canada, April 14-16, 2014, Conference Track Proceedings}, 2014.

\bibitem{jang2016categorical}
\BIBentryALTinterwordspacing
E.~Jang, S.~Gu, and B.~Poole, ``Categorical reparameterization with
  gumbel-softmax,'' 2016, cite arxiv:1611.01144. [Online]. Available:
  \url{http://arxiv.org/abs/1611.01144}
\BIBentrySTDinterwordspacing

\bibitem{rasmussen2003gaussian}
C.~E. Rasmussen and C.~K.~I. Williams, \emph{Gaussian processes for machine
  learning.}\hskip 1em plus 0.5em minus 0.4em\relax MIT Press, 2006.

\bibitem{friedman2001elements}
T.~Hastie, R.~Tibshirani, and J.~Friedman, \emph{The Elements of Statistical
  Learning}, ser. Springer Series in Statistics.\hskip 1em plus 0.5em minus
  0.4em\relax New York, NY, USA: Springer New York Inc., 2001.

\bibitem{white2011multiple}
I.~R. White, P.~Royston, and A.~M. Wood, ``Multiple imputation using chained
  equations: issues and guidance for practice,'' \emph{Statistics in medicine},
  vol.~30, no.~4, pp. 377--399, 2011.

\bibitem{autoimpute}
\BIBentryALTinterwordspacing
``Autoimpute.'' [Online]. Available:
  \url{https://kearnz.github.io/autoimpute-tutorials/}
\BIBentrySTDinterwordspacing

\bibitem{fancyimpute}
\BIBentryALTinterwordspacing
A.~Rubinsteyn and S.~Feldman, ``fancyimpute: An imputation library for
  python.'' [Online]. Available: \url{https://github.com/iskandr/fancyimpute}
\BIBentrySTDinterwordspacing

\bibitem{mice}
F.~Pedregosa, G.~Varoquaux, A.~Gramfort, V.~Michel, B.~Thirion, O.~Grisel,
  M.~Blondel, P.~Prettenhofer, R.~Weiss, V.~Dubourg, J.~Vanderplas, A.~Passos,
  D.~Cournapeau, M.~Brucher, M.~Perrot, and E.~Duchesnay, ``Scikit-learn:
  Machine learning in {P}ython,'' \emph{Journal of Machine Learning Research},
  vol.~12, pp. 2825--2830, 2011.

\end{thebibliography}

\end{document}